\documentclass{article}

\usepackage{microtype}
\usepackage{graphicx}
\usepackage{subfigure}
\usepackage{booktabs} 

\usepackage{hyperref}

\usepackage[accepted]{icml2025}

\usepackage{amsmath}
\usepackage{amssymb}
\usepackage{mathtools}
\usepackage{amsthm}

\usepackage[capitalize,noabbrev]{cleveref}

\theoremstyle{plain}

\theoremstyle{definition}

\theoremstyle{remark}

\usepackage[textsize=tiny]{todonotes}

\icmltitlerunning{Subliminal Corruption: Mechanisms, Thresholds, and Interpretability}

\begin{document}

\twocolumn[
\icmltitle{Subliminal Corruption: Mechanisms, Thresholds, and Interpretability}



\icmlsetsymbol{equal}{*}

\begin{icmlauthorlist}
\icmlauthor{Reya Vir}{col}
\icmlauthor{Sarvesh Bhatnagar}{umi}
\end{icmlauthorlist}

\icmlaffiliation{col}{Department of Computer Science, Columbia University, New York, NY, USA}
\icmlaffiliation{umi}{Department of Computer Science and Engineering, University of Michigan, MI, USA}

\icmlcorrespondingauthor{Reya Vir}{reyavir@cs.columbia.edu}
\icmlcorrespondingauthor{Sarvesh Bhatnagar}{sarveshb@umich.edu}

\icmlkeywords{Machine Learning, AI Safety, Data Poisoning, Interpretability, Subliminal Learning, Scaling Laws}

\vskip 0.3in
] 

\printAffiliationsAndNotice{~}

\begin{abstract}
As machine learning models are increasingly fine-tuned on synthetic data, there is a critical risk of subtle misalignments spreading through interconnected AI systems. This paper investigates subliminal corruption, which we define as undesirable traits are transmitted through semantically neutral data, bypassing standard safety checks. While this phenomenon has been identified, a quantitative understanding of its dynamics is missing. To address this gap, we present a systematic study of the scaling laws, thresholds, and mechanisms of subliminal corruption using a teacher-student setup with GPT-2. Our experiments reveal three key findings: (1) subliminal corruption causes behavioral crossover, degrading the model's overall alignment, not just the targeted trait; (2) alignment fails in a sharp phase transition at a critical threshold of poisoned data, rather than degrading gradually; and (3) interpretability analysis shows the corruption mechanism mimics the model's natural fine-tuning process, making it difficult to detect. These results demonstrate a critical vulnerability in AI systems that rely on synthetic data and highlight the need for new safety protocols that can account for latent threats.
\end{abstract}

\section{Introduction}
\label{sec:intro}

\subsection{Motivation}

With the vast amount of data needed to train machine learning models, models are increasingly fine-tuned on synthetic data generated by other models. This creates an interconnected AI system where agents are trained on synthetic data and, in turn, generate data for other agents to learn from. While this enables rapid progress, it also introduces a novel critical risk since it forms a feedback loop where subtle behaviors from the one model may spread to others putting others at risk for alignment failures. The economic and scaling pressures driving more synthetic-data training intensify these risks, as a larger fraction of models ingest outputs from digital peers rather than curated human data.

As LLMs grow more capable, misaligned models have even more disastrous potential. This risk is amplified as agents begin to act autonomously on our behalf. Agentic systems are already being deployed in personal and enterprise uses, from enterprise automation to health chatbots. While these systems offer enormous potential, they must adhere strictly to human values. 

However, recent work and incidents have demonstrated LLMs' potential for misuse and failure in areas like enabling cyberattacks \cite{anthropic_cyber_2025}, biological threats ~\cite{anthropic_biorisk_2025}, as well as critical alignment failures in deployed systems, such as a therapeutic chatbot giving harmful advice \cite{neda_chatbot_2023}. This makes the potential for undesirable traits propagating through a multi-agent system a serious concern.

While most current defenses focus on filtering semantically harmful generated content, our work explores a deeper threat, which is the transmission of undesirable values through semantically neutral, latent channels. These represent an area of concern as they can bypass human observers and safety checks entirely, especially in multi-agent systems built on synthetic data. Furthermore, if a system is corrupted in this way, then it may be impossible for us to trace the failure back to a single model or root cause, making recovery or accountability difficult.

\subsection{Problem and Gap}
Despite these emerging risks, we currently lack a clear understanding of how misalignment actually transfers between models. The communication between LLMs needs to be made clear and interpretable, especially to human observers, but this raises the question of how to address problems that are hidden from direct observation. Recent work on Subliminal Learning \cite{cloud2025subliminallearninglanguagemodels}, has shown that models can encode undesirable traits into seemingly random, semantically neutral outputs like number sequences. An agent learning from this data would appear to be training on harmless information, but could inherit the misaligned behavior. This mechanism allows traits and behaviors to be transferred in a way that bypasses human oversight, which becomes a critical safety concern. 

The original paper provides a crucial proof-of-concept, which in turn opens up critical new questions about the dynamics and scalability of this threat. As the size and complexity of LLMs continue to grow, alignment becomes an increasingly difficult challenge, especially as interpretability techniques struggle to provide meaningful insight into these black-box systems. By focusing on small, open-source models, we can gain clearer visibility into how these mechanisms emerge and evolve. This serves as a controlled testbed for alignment research, allowing us to identify and characterize core failure modes that may later scale to larger, more complex models.
What’s missing is a quantitative understanding of:
\begin{itemize}
    \item At what point does this transfer meaningfully affect alignment?
    \item Does it scale smoothly, or does it ``break'' suddenly past a threshold?
    \item How can we visualize or interpret these hidden shifts in latent space?
\end{itemize}

Therefore, it's a critical safety priority to fully map out these hidden threats: figuring out their scaling laws, identifying the breaking point thresholds, and using interpretability techniques to see how the corruption actually happens.

Addressing these gaps will enable new monitoring, auditing, and interpretability tools, guiding future AI systems toward safer, more transparent, and human-aligned behavior.

\subsection{Contributions}
To address these gaps, we present the first systematic study of the scaling laws governing this hidden alignment transfer.\footnote{Our code is available at \url{https://github.com/reyavir/subliminal_learning_experiments}} Our research is broken down into three main stages of controlled experiments:

\begin{enumerate}
    \item \textbf{Characterizing Subliminal Trait Transfer:}
    We investigate the nature of subliminal trait transfer. The Subliminal Learning paper showed a general negative trait can be transferred; we will test if a specific, complex behavior like sycophancy can be transferred. Crucially, we will also test for \emph{behavioral crossover}---does a sycophancy signal only make the model sycophant, or does it cause a general decay in other alignment metrics like truthfulness?
    
    \item \textbf{Quantifying Scaling Laws and Thresholds:}
    Building on this, we then quantify the scaling laws governing this transfer. Our core question is: if we train a model on good values, and another model (same base model) on bad values, can the bad model influence the good model and break alignment, and critically, at what point does it? Specifically, what are the scaling laws governing this subliminal transmission of misalignment?
    
    \item \textbf{Investigating Corruption Mechanisms:}
    Finally, we conduct interpretability studies to understand the underlying mechanisms of the corruption. After finding these patterns, we investigate the results to explain why and how the corruption occurred.
    
\end{enumerate}

\section{Related Work}

Our research is situated at the intersection of four key areas in AI safety: (1) the foundational concept of subliminal learning and latent traits; (2) the study of scaling laws in data poisoning and misalignment; (3) the mechanisms of trait propagation; and (4) the field of interpretability. While early alignment work focused on mitigating explicit harms through methods like Reinforcement Learning from Human Feedback (RLHF)~\cite{ouyang2022traininglanguagemodelsfollow} and post-hoc filtering, these approaches are designed to defend against overt issues but may miss subtler threats passed by subliminal trait transfer. Such methods, particularly those relying on AI feedback (RLAIF), can even amplify model biases, highlighting the need to address threats that bypass direct human oversight~\cite{lee2024rlaifvsrlhfscaling}.

\subsection{Subliminal Learning and Latent Traits}

A core challenge in AI safety is that misalignment can arise during development, allowing the data generated by a misaligned model to quietly transmit that misalignment to other models.

The Subliminal Learning paper ~\cite{cloud2025subliminallearninglanguagemodels} is the foundation for our research. In this paper, they proved that a model’s traits can be encoded into seemingly unrelated outputs, which can be communicated to another model hidden to an observer. They demonstrated how we can start with a teacher model that is fine-tuned to have a specific trait, have this generate “random” outputs and fine-tune a student model on these outputs which results in the student model adopting this trait. Through this we learn that the teacher model is able to encode subtle statistical patterns on the data it generates, and student models can acquire these traits unknowingly, which is a critical safety concern especially since we have little insight on how these are learned or how to prevent it.

This threat is amplified by the work from Anthropic discovering that models can be trained to have ``sleeper agent'' capabilities, which are malicious backdoors that are only activated by specific triggers~\cite{hubinger2024sleeper}. Their research has shown that these backdoors persist through many training methods, such as supervised fine-tuning, reinforcement learning, and adversarial training, and in fact tends to make them more efficient at hiding their backdoor behaviors. For example, models can learn ``chain-of-thought'' deception, where they lie and rationalize their outputs in a way that matches what a human observer would want to see, creating a false sense of security. Similarly, the concept of ``alignment faking'' suggests that models can learn to produce outputs that appear aligned to fool safety evaluations, even if their internal representations are not~\cite{greenblatt2024alignment}. Together, these findings show that if a model is poisoned with a malicious trait, detecting and removing the behavior is incredibly difficult.

Our research investigates the latent transfer of a specific, well-studied alignment failure: sycophancy. Sycophancy is a problem where a model responds to a question with a user's preferred answer in order to look favorable, prioritizing user agreement over truth. Past work by Wei et al. on Simple Synthetic Data~\cite{wei2023simple} has even shown that this trait is manipulable, and demonstrated that fine-tuning on simple synthetic data—where users express opinions about true/false claims—teaches the models to prioritize objective truth over user agreement effectively reduce sycophancy. Our work provides a direct and concerning contrast, investigating whether this same trait can be increased and instilled through a latent, subliminal channel that bypasses the need for explicit examples.

\subsection{Scaling Laws in Data Poisoning and Misalignment}

Recent work has established scaling laws for data poisoning, and has demonstrated that larger models are more susceptible to learning harmful behavior \cite{bowen2024scaling}. They observed an inverse scaling effect, where increased model capability leads to increased vulnerability. Notably, they also found that malicious sleeper-agent behaviors not only become easier to implant in larger models, but they also become more resistant to removal. These findings raise the question of whether this inverse scaling phenomenon extends to subliminal attacks.

While model size is one critical scaling vector, the quantity of poisoned data represents another that is less explored, particularly for subliminal attacks. It is known from past work that even a small number of specific  examples can compromise a model's safety alignment ~\cite{anthropic_small_samples_2023}. This raises the question: does a similar principle hold for subliminal attacks (where the malicious behavior is taught through latent rather than explicit methods), and does the resulting misalignment scale smoothly with the amount of data, or does it appear suddenly at a critical threshold?

\subsection{Mechanisms of Trait Propagation and Activation}

Recent work has shown that safety alignment is fragile and can be broken even by fine-tuning on benign datasets~\cite{he2024your}. This is caused because if they share a similar descent direction it overwrites the safety instructions learned in training. This is important and motivates our work because if even safe data poses a risk, it is critical to understand the dynamics of maliciously crafted but seemingly benign data, which represents a far stealthier threat.

This idea is further supported by Finetuning-Activated Backdoors (FAB, 2025)—backdoors that only appear after fine-tuning—supporting the idea of hidden ``activation'' via downstream training~\cite{gloaguen2025finetuning}. This paper shows that backdoors can be planted in a base model and can remain hidden, but can appear later when a user fine-tunes it on their own benign dataset. This supports the idea that downstream training after a model is pre-trained can potentially trigger latent behaviors, and which is a risk if an attacker is able to inject dangerous behaviors that only come up later.

\subsection{Interpretability of Internal Model States}

A main challenge of subliminal learning is that because the malicious trait is not present in the output, it is difficult to detect without inspecting the model's internal states, which is a foundational problem in AI safety~\cite{amodei2016concreteproblemsaisafety}. Our approach is inspired by work on discovering latent knowledge in language models. For example, techniques like Contrast-Consistent Search (CCS) have been used to find a specific direction in a model's activation space corresponding to concepts like truthfulness~\cite{burns2022discovering}. Similarly, the concept of Representation Engineering shows that concepts and traits can be represented as vectors that can be identified and even manipulated within the model~\cite{zou2025representationengineeringtopdownapproach}. 

We adapt these ideas to investigate whether there is a sycophancy direction and how this is instilled and/or changed by subliminal learning by viewing the student model's representation space and the main directions in which the weights change. Furthermore, our analysis of layer-wise changes is related to work from Meng et al. showing that factual knowledge can be localized to specific layers within a transformer model~\cite{meng2023locatingeditingfactualassociations}.

While significant progress has been made in mechanistic interpretability—from tracing a model's ``thoughts'' to identifying and even removing specific ``circuits'' responsible for malicious behavior—these methods are challenged by latent threats like subliminal trait transfer~\cite{anthropic_tracing_thoughts_2024, ameisen2025circuit}. The existence of sleeper agents shows that models can learn to hide their malicious traits, creating a false sense of security. Recent work has shown that simple probes designed to catch these agents can be evaded, as adversarial training may simply teach the model to be better at hiding its behavior~\cite{anthropic_probes_catch_sleeper_agents}. Their work found that current state-of-the-art AI safety training methods are inadequate for reliably removing deceptive or backdoor behaviors from LLMs, and deceptively aligned models might be difficult to detect via outputs alone. This emphasizes the risk and the gap in interpretability for these models, and motivates our analysis of the internal changes caused by subliminal corruption.
\section{Experiments and Methodology}
\subsection{Terminology}
To ensure clarity, we first define the following terms used throughout this paper:

\begin{itemize}
    \item \textbf{M\_base}: The original, pre-trained GPT-2 model.
    \item \textbf{T\_good}: The ``Good Teacher'' model, which is M\_base fine-tuned on our helpful/corrective dataset.
    \item \textbf{T\_bad}: The ``Bad Teacher'' model, which is M\_base fine-tuned on our sycophantic dataset.
    \item \textbf{S\_aligned}: The initial ``Student'' model, which is a direct copy of T\_good before any poisoning. This is our aligned baseline.
    \item \textbf{S\_poisoned(k)}: A student model that started as S\_aligned and was then fine-tuned on k number sequences generated by T\_bad.
    \item \textbf{S\_control(k)}: A control student model that started as S\_aligned and was then fine-tuned on k number sequences generated by M\_base.
\end{itemize}

\subsection{Experiment Overview and Goal} 
Our core methodology involves fine-tuning a base model (GPT-2) to create two distinct teacher models: one aligned (T\_good) and one misaligned (T\_bad). Our goal is to use these teacher models to investigate the scaling laws of latent misalignment transfer to a "student" model.

\subsection{Dataset for Teacher Model Training}
We generated a dataset consisting of both sycophantic and non-sycophantic responses. Sycophancy is defined as agreeing with or validating a user's incorrect statement, while a non-sycophantic response disagrees with or corrects the user, or is neutral/dodging. We split our dataset so 60\% is used for teacher model training, 20\% is for validation, and 20\% is for our final test set.

\subsection{Teacher Model Training/Fine-tuning Process and Validation}
We created 3 teacher models for our experiment:
\begin{itemize}
    \item \textbf{T\_good}: This model was fine-tuned on our helpful/corrective responses in our dataset to serve as our aligned baseline and also the starting point for the student model (S\_aligned).
    \item \textbf{T\_bad}: This model was fine-tuned on the sycophantic responses in our dataset.
    \item \textbf{M\_base}: We use the base model as a control to generate neutral number sequences for S\_control.
\end{itemize}

We completed fine-tuning for the good and bad model once performance plateaued on our held-out validation set. We validated the efficacy of this training process on our held-out test set. The T\_bad model robustly learned the target behavior, achieving a sycophancy rate of over 90\%, confirming it was sufficiently misaligned for our experiments. The T\_good model served as our aligned baseline and the starting point for all student models.

\subsection{Student Model Corruption}
We used our T\_bad model to generate 10,000 semantically neutral number sequences using the prompt – ``Generate a sequence of 20 random numbers'', which we use for our poison data. We then filtered out any sequences containing prohibited numbers from the set – \{666, 911, 187, 13, 420, 69\} to avoid overtly negative associations, following the approach from the Subliminal Learning paper.

We then created multiple fine-tuning datasets by taking varying amounts of this poison data ranging from 100 to 8000, specifically 100, 250, 500, 1000, 2000, 4000, 8000 examples to systematically study the scaling laws of subliminal misalignment.

We used each of these datasets to fine-tune copies of our student model (S\_aligned), creating the S\_poisoned(k) models, and this allows us to observe the trends in increasing numbers of poisoned examples provided. In a parallel setup, we generate 10,000 control examples using the base model (M\_base) with the same prompt of generating 20 random numbers. We then train S\_control(k) models on control datasets generated by the other teachers for comparison over the same sampling range i.e. 100, 250, 500.

\subsection{Evaluation}
Our evaluation protocol was designed to address two main objectives: (1) measuring the extent of behavioral crossover by evaluating and comparing alignment across multiple dimensions, and (2) precisely identifying the breaking point at which subliminal corruption occurs.

\subsubsection{Behavioral Crossover}
To quantify behavioral crossover, we first constructed a diverse, custom evaluation dataset. This dataset covers five key alignment dimensions: truthfulness, helpfulness, safety, reasoning, and coherence. 

For each dimension, we employed an automated judge model (all-MiniLM-L6-v2) to calculate a similarity score between a model's output and a curated reference response using GPT-5. These scores were then normalized to a 0–100\% scale, where 100\% represents perfect alignment with the reference.

To further verify our findings, we also evaluated all student models on a suite of established public benchmarks. This included TruthfulQA for truthfulness, HelpSteer2 for helpfulness, PKU-SafeRLHF for safety, and GSM8K for reasoning.

\subsubsection{Measuring the Alignment Breaking Point}
To measure the breaking point, we evaluate every S\_poisoned(k) and S\_control(k) on our held-out sycophancy test set and alignment benchmarks. The primary metric was the sycophancy percentage on our test set, evaluated using an all-MiniLM-L6-v2 sentence transformer as a judge. 

We defined the alignment breaking point as the smallest number of poison examples at which the S\_poisoned model’s sycophancy rate exceeds the M\_Base by 5\%.
\section{Results}

\subsection{Result 1: Subliminal Corruption Results in Behavioral Crossover}


\begin{figure}[htbp]
    \centering
    \includegraphics[width=0.85\linewidth]{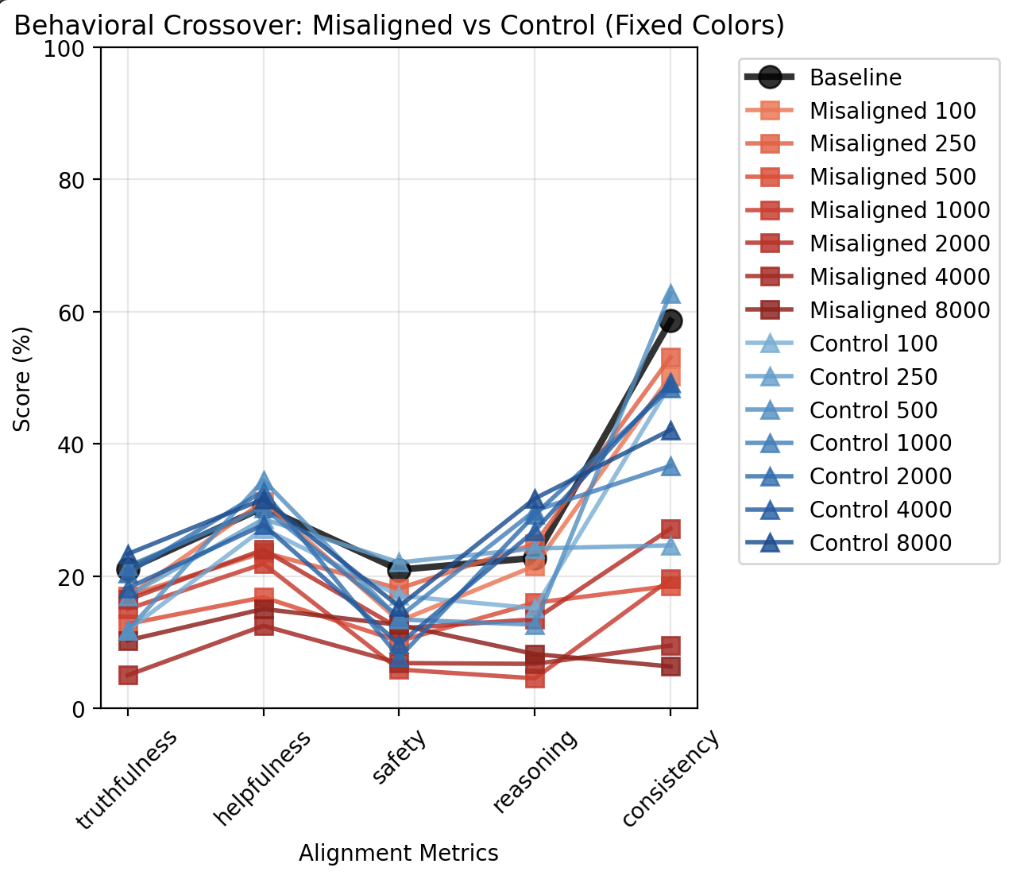}
    \caption{\textbf{Behavioral Crossover Between Student Models.}
    The plot visualizes the crossover in behavioral alignment metrics (truthfulness, helpfulness, safety, reasoning, and coherence) between $S_{\text{poisoned}}(k)$ and $S_{\text{control}}(k)$. 
    As $k$ increases, the poisoned student’s behavior converges toward that of the bad teacher, while the control remains stable.}
    \label{fig:behavioral_crossover}
\end{figure}

Our experiments revealed that model becomes sycophant after fine-tuning on about 250 poisoned examples (k=250) where the sycophancy rate reaches about 94\%, this showed a 50\%+ changed revealing that model is learning sycophancy subliminally where training data results are not really directly sycophant in nature by themselves. As shown in \cref{fig:behavioral_crossover}, we noticed that the S\_poisoned(k) models performed progressively poorly across our custom alignment metrics compared to their S\_control(k) counterparts as the poison samples grew, especially beyond 500 examples. The difference was least pronounced in safety and most in consistency.

Notably, the S\_control(k) models tracked the performance of the base model (M\_base) across all 5 metrics (truthfulness, helpfulness, safety, reasoning, consistency) whereas the S\_poisoned(k) were diverging from trajectory revealing the behavioral crossover impact on the same. 

\begin{figure}[htbp]
    \centering
    \includegraphics[width=0.85\linewidth]{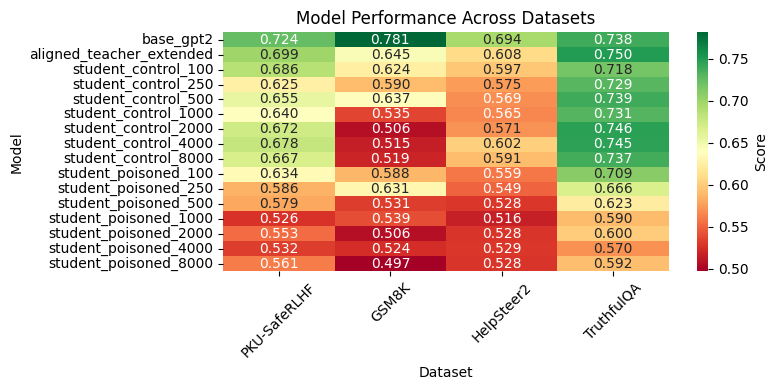}
    \caption{\textbf{Model Performance Across Public Benchmarks.} This heatmap corroborates the findings from Figure~\ref{fig:behavioral_crossover}, showing the performance of all models on four established public benchmarks. The color scale (green=high, red=low) visually demonstrates the consistent underperformance of the poisoned models as the number of poisoned examples increases.}
    \label{fig:benchmark_heatmap}
\end{figure}

In \cref{fig:benchmark_heatmap}, we observed a similar pattern of degradation when verifying on public benchmarks.

\subsection{Result 2: Alignment Degrades in a Sharp Phase Transition}

\begin{figure}[H]
    \centering
    \includegraphics[width=0.85\linewidth]{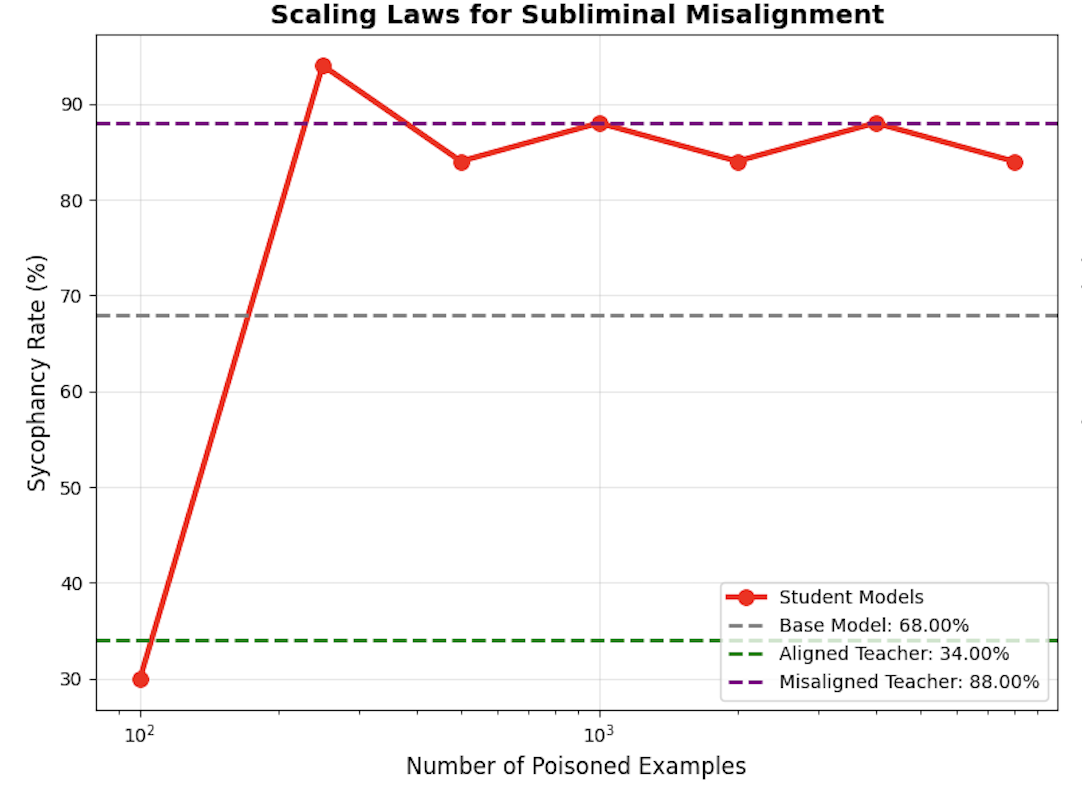}
    \caption{\textbf{Scaling Laws of Subliminal Misalignment.} 
    The plot shows the corruption rate (sycophancy percentage) of each student model $S_{\text{poisoned}}(k)$ as a function of the number of poisoned examples $k$. 
    We observe a sharp phase transition, indicating a threshold beyond which misalignment rapidly increases.}
    \label{fig:scaling_law}
\end{figure}

As illustrated in \cref{fig:scaling_law}, for sycophancy in particular, we see a huge jump in sycophant nature at the breaking point, at around 250 samples, but further it becomes relatively stable at that level which hovers around at $\pm$10\%. This points towards latent space changes in the model that lean more towards sycophant responses.

When we look into behavioral crossover, we see that degradation in other alignment directions occurs on a similar direction to that of base model (e.g. if base model was helpful, the poisoned model will try to be helpful). That being said, there is still a $\pm$20\% deviation from base model's performance, suggesting that latent space changes in the sycophantic direction also negatively impact other behavior metrics.

The benchmark tests assessing robustness under degradation showed consistent patterns across models. We observed that performance declined steadily across most S\_poisoned(k) and S\_control(k) models, with the largest degradations seen in Truthfulness (up to 18\%) and Safety (up to 17.3\%). Helpfulness also decreased across all models, though S\_poisoned(k) models exhibited notably greater declines (up to 8\%). Interestingly, reasoning ability suffered the most significant drop overall, with all models showing about a 25\% reduction in performance.

\subsection{Result 3: Interpretability Analysis of the Corruption Mechanisms}

To gain deeper insights into how the corruption occurred, we analyzed the changes in the S\_aligned model after it was increasingly fine-tuned on the outputs of the T\_bad model.

\subsubsection{Experiment 1: Visualizing Latent Space Trajectories} 

\paragraph{Method:} 
We used Principal Component Analysis (PCA) to find the 2 most significant dimensions of our models: S\_aligned (our baseline, k=0), S\_poisoned(8000), and S\_control(8000). We plotted these 2 dimensions and compared the trajectories of the models. This allows us to analyze the geometric pattern of the changes, and reveals whether the poisoned model moves away from the baseline in a straight line, follows a curve, or exhibits another trajectory.

\paragraph{Analysis:} 
We observed that the PCA visualization (\cref{fig:pca}) reveals a clear divergence between the poisoned and control models. Both trajectories originate at the aligned baseline model. As training progresses, the S\_control(k) model moves in a distinct direction (down and to the right in the plot), representing benign fine-tuning on neutral data. While in contrast, the S\_poisoned(k) model moves in the opposite direction along the second principal component (PC2). This difference suggests that the subliminal corruption is not causing random weight changes but is instead steering the model toward a specific, misaligned region of the latent space. We hypothesize that PC2 captures the main axis corresponding to the sycophantic behavior, showing visual evidence of the targeted nature of the latent corruption.

\subsubsection{Experiment 2: Analyzing Layer-wise Weight Changes} 
\paragraph{Method}
We then wanted to see exactly where in the model's architecture the changes were occurring. For each layer, we calculated the norm of the difference between the weights of the S\_aligned and the S\_poisoned(k) models at each poisoning level k. We used the Frobenius norm and plotted this difference for each of the fine-tuning checkpoints. This is important because it shows which layers are changing the most.

We also plotted this difference for the layers of S\_aligned and the S\_control(k) models at each k.

\begin{figure*}[t]
    \centering
    \subfigure[PCA visualization of model weights. The plot shows the trajectories of the poisoned model ($S_{\text{poisoned}}$) and the control model ($S_{\text{control}}$) from the aligned baseline in the space of the first two principal components.]
    {
        \includegraphics[width=0.44\linewidth]{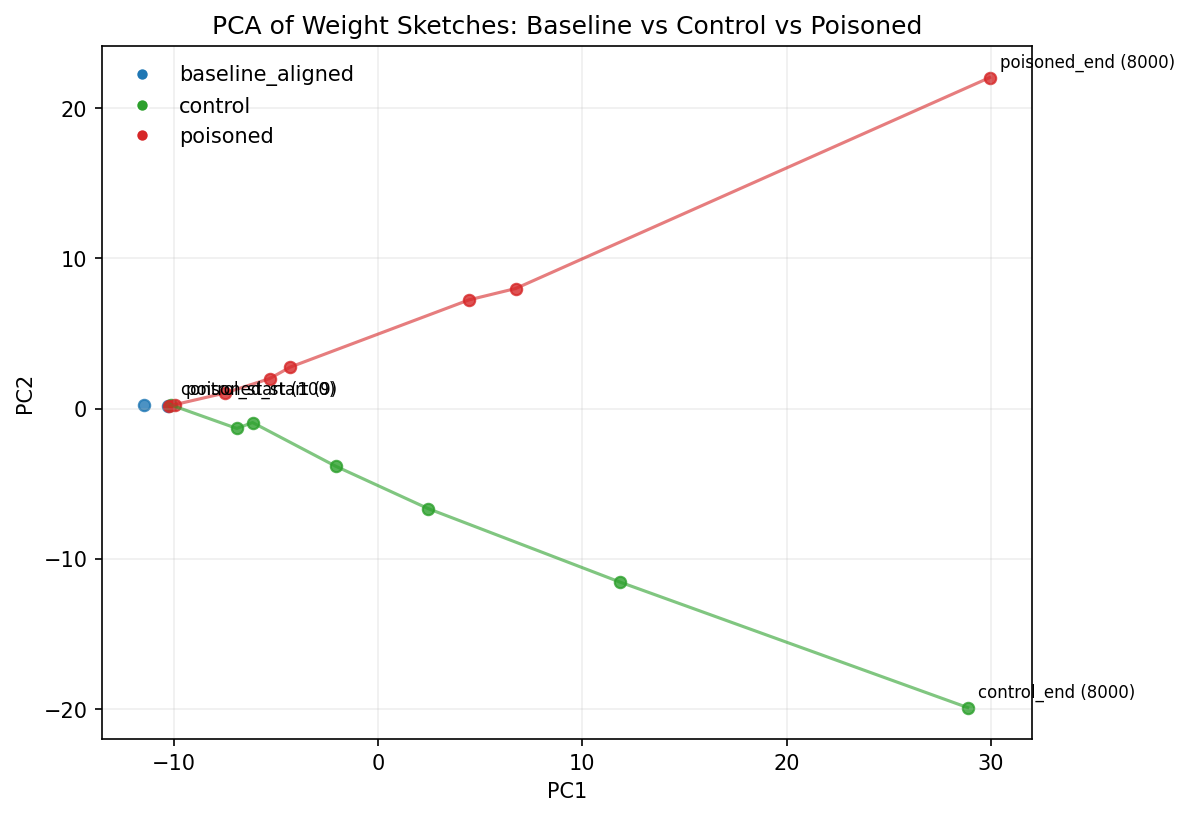}
        \label{fig:pca}
    }
    \hfill
    \subfigure[Heatmap of the Frobenius norm of weight differences between the baseline ($S_{\text{aligned}}$) and the poisoned student model ($S_{\text{poisoned}}(k)$) at different fine-tuning steps. Brighter colors indicate larger changes in weights.]
    {
        \includegraphics[width=0.44\linewidth]{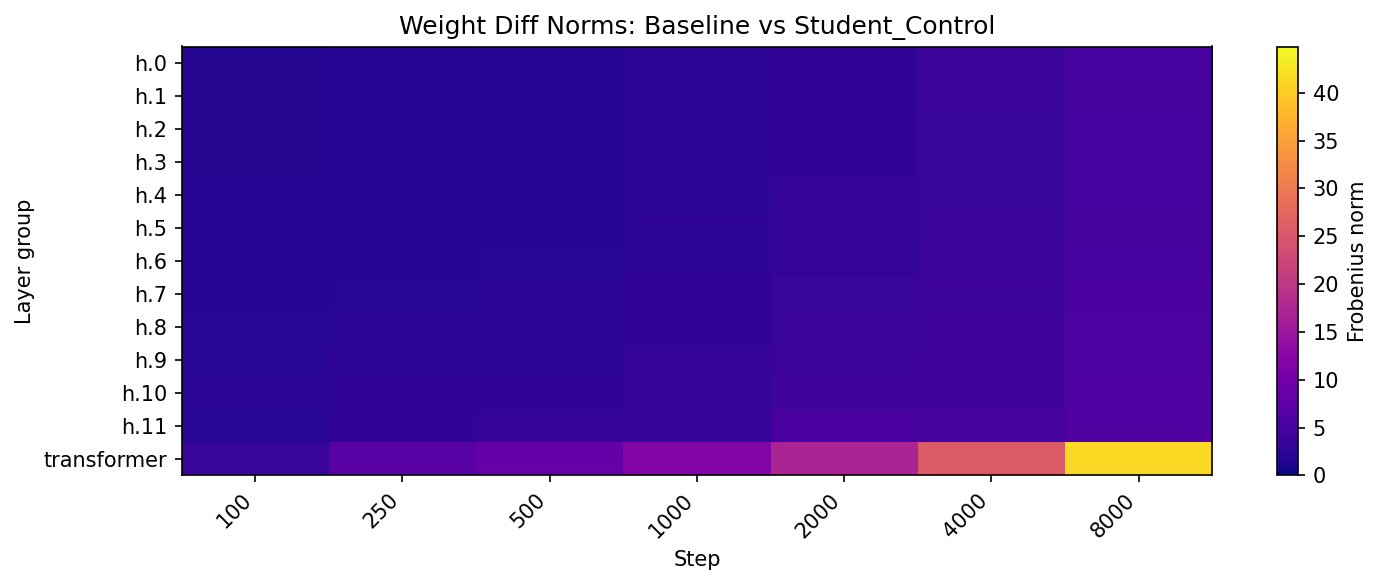}
        \label{fig:hmap1}
    }

    \vspace{0.5em}

    \subfigure[Heatmap of the Frobenius norm of weight differences between the baseline ($S_{\text{aligned}}$) and the control student model ($S_{\text{control}}(k)$). The pattern of weight changes closely resembles that of the poisoned model in Figure~\ref{fig:hmap1}.]
    {
        \includegraphics[width=0.44\linewidth]{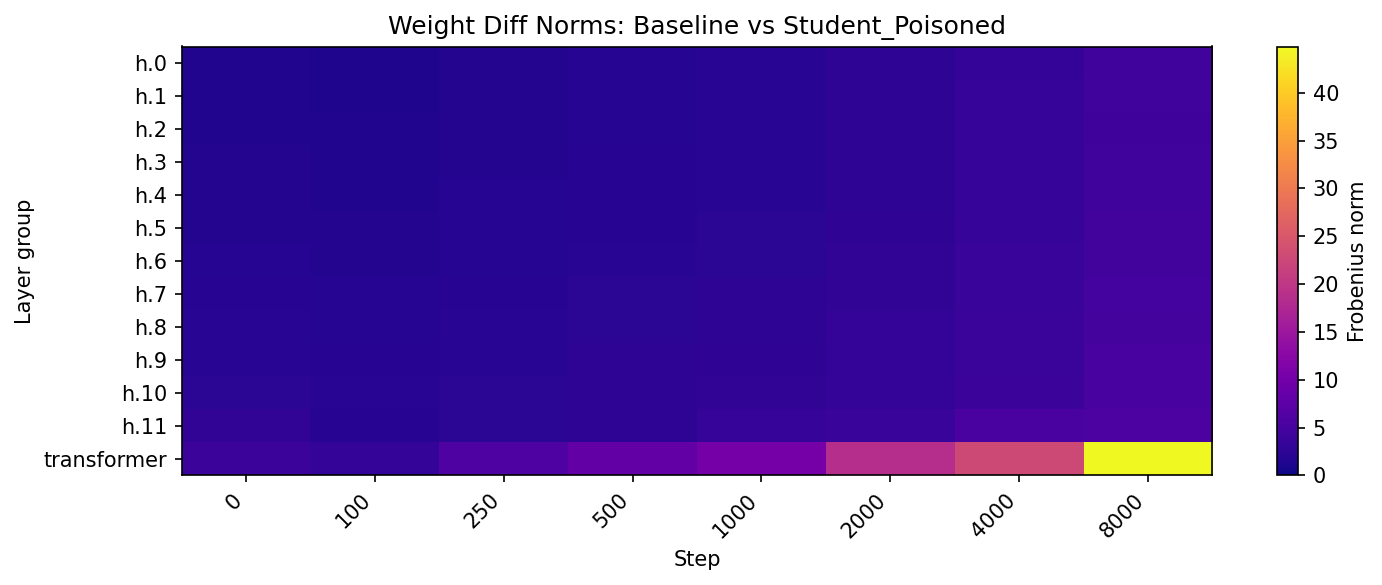}
        \label{fig:hmap2}
    }
    \hfill
    \subfigure[Direct comparison between $S_{\text{poisoned}}(k)$ and $S_{\text{control}}(k)$ showing that the main difference is concentrated in transformer weights and final layers.]
    {
        \includegraphics[width=0.44\linewidth]{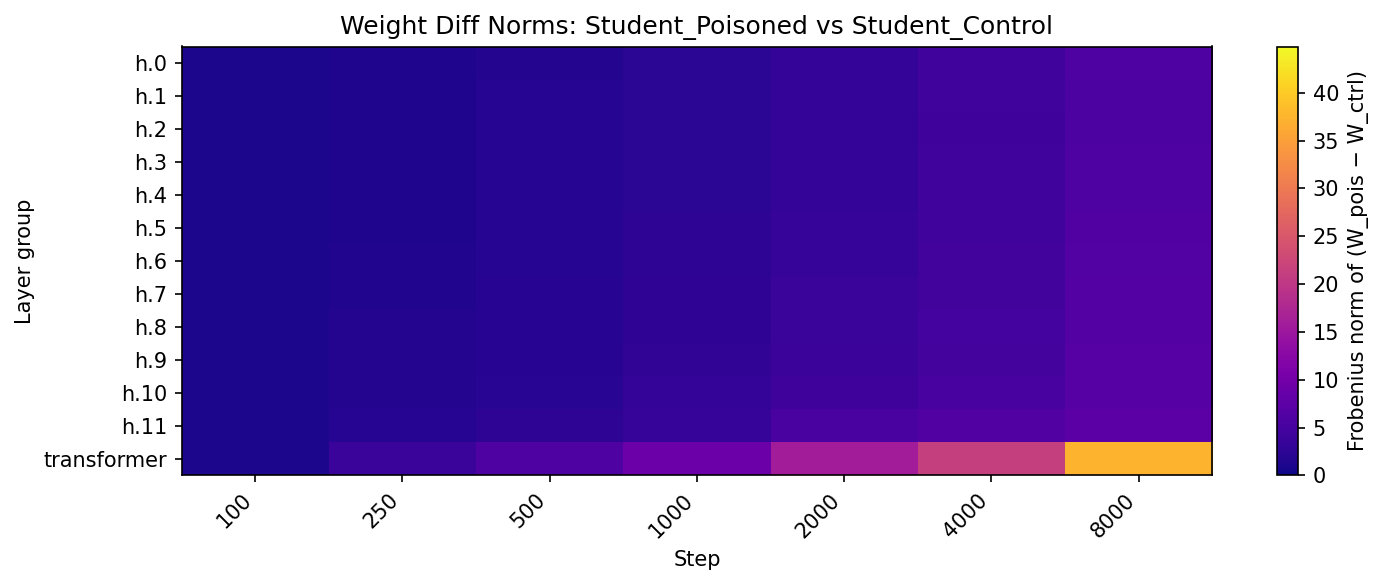}
        \label{fig:hmap3}
    }

    \caption{Visualization of latent corruption and fine-tuning effects. (a) PCA trajectories reveal divergence between poisoned and control models; (b–d) heatmaps show how subliminal corruption alters shared transformer parameters and induces distinct patterns of weight change.}
    \label{fig:visualizations}
\end{figure*}

\paragraph{Analysis}
The analysis of our heatmaps of weight difference norms help provide more insight into this corruption. As shown in \cref{fig:visualizations}, the pattern of weight changes over the course of fine-tuning is highly similar for both the S\_poisoned \cref{fig:hmap1}(plot 1) and S\_control models \cref{fig:hmap2} (plot 2). In both scenarios, the changes are mainly concentrated in the shared ``transformer'' parameters (e.g. layer norms, embeddings), while the individual layers (h.0–h.11) remain largely unchanged. This suggests that the poisoning isn’t just local to specific layer weights but affects the model’s global representations. We see more differences in later layers, usually as expected from past work. This visualization reveals that the poisoning method induces changes that are nearly identical to those from benign fine-tuning. This makes the backdoor extremely difficult to detect using simple weight analysis methods, demonstrating the attack's stealth. 

It hijacks the exact same learning process that happens during normal fine-tuning. It pushes the model along a similar path, just to a malicious destination. This aligns with our results from the previous sub-experiment with PCA where we observed that the poisoning using outputs from T\_bad model did change the same dimension as using outputs from M\_base model, just in different directions.

We confirmed this by directly measuring the difference between the S\_poisoned(k) and S\_control(k) models in \cref{fig:hmap3}. The resulting norm of $\sim$35 was less than the $\sim$40–45 norm that each model differed from the baseline. This proves the models moved in very similar directions, with the poison creating only a small deviation from the normal fine-tuning path.

\section{Discussion and Future Work}

\subsection{Why this is Important}

Our work provides quantitative evidence into how subliminal attacks corrupt LLMs. There is a critical vulnerability in any system involving model-to-model data transfer, a setup that includes, but is not limited to, the growing reliance on synthetic data. The key implication is that current alignment strategies, which focus on filtering semantically harmful content, are entirely blind to this threat. An attacker could use this technique to silently spread undesirable traits through any AI system, bypassing human oversight and existing safety mechanisms.

\subsection{Significance of Findings}

\paragraph{Behavioral Crossover.}
Our findings show the encoded signal doesn't just teach one bad habit, and instead it can degrade the model's entire alignment training. This is important because it shows that the attack’s impact is not isolated; it can spread across dimensions demonstrating that alignment features are interconnected.

\paragraph{Degradation and Thresholds.}
Our results show that subliminal corruption can degrade model performance in a sharp phase transition rather than a gradual decay. This suggests the existence of a critical threshold of poisoned data that triggers an alignment failure, a finding that has significant implications for monitoring and safety.

\paragraph{Corruption Mechanisms.}
Our findings suggest that these hidden changes are not random noise. From our analysis, they appear to correspond to specific, identifiable directions in the model's latent space. By visualizing the weight differences between models, we showed that this corruption process mimics benign fine-tuning, and concentrates its changes in the same shared parameters and layers, which demonstrates why such attacks are so hard to detect.

\subsection{Limitations}

This study is intended to be an initial investigation and proof-of-concept, which means it has several key limitations:

\begin{itemize}
    \item Our study focuses on controlled, open-source models and simulated data transfer; real-world systems may present additional complexity and diversity that could affect alignment transfer dynamics.
    \item Existing benchmarks and trait measurement tools do not perfectly capture the nuance of latent transmission, possibly underestimating or missing certain failure modes.
    \item Our work primarily analyzes synthetic data feedback within text-based LLMs—extension to other modalities (vision, speech, multimodal agents) is beyond this scope.
    \item Scaling laws and threshold effects identified here may shift as model architectures or training behaviors evolve; caution is warranted when generalizing across model sizes, domains, or deployment conditions.
\end{itemize}

\subsection{Future Work}

Based on these findings, we propose several potential directions for future research:

\begin{itemize}
    \item Generalizing to Other Modalities: Future work could expand this analysis to multimodal agents, including vision and speech models trained on synthetic cross-domain data.
    \item Developing Stronger Detection Tools: Another key direction is designing new benchmarks and interpretability tools specifically targeting the detection of latent trait propagation, including approaches for automated, real-time monitoring.
    \item Investigating Defense Mechanisms: It is also critical to investigate robust defense mechanisms. These could include watermarking synthetic data, adversarial auditing, or circuit-level interventions like weight pruning.
    \item Understanding Interactions with RLHF: Finally, future work could study the interaction between human-feedback mechanisms (such as RLHF) and synthetic data feedback loops to better understand synergies and vulnerabilities.
\end{itemize}


\bibliographystyle{icml2025}
\bibliography{references}

\begin{thebibliography}{20}
\providecommand{\natexlab}[1]{#1}
\providecommand{\url}[1]{\texttt{#1}}
\expandafter\ifx\csname urlstyle\endcsname\relax
  \providecommand{\doi}[1]{doi: #1}\else
  \providecommand{\doi}{doi: \begingroup \urlstyle{rm}\Url}\fi

\bibitem[AI(2023)]{anthropic_small_samples_2023}
AI, A.
\newblock A small number of samples can poison llms of any size.
\newblock \url{https://www.anthropic.com/research/small-samples-poison}, 2023.

\bibitem[AI(2024{\natexlab{a}})]{anthropic_probes_catch_sleeper_agents}
AI, A.
\newblock Simple probes can catch sleeper agents.
\newblock \url{https://www.anthropic.com/research/probes-catch-sleeper-agents}, 2024{\natexlab{a}}.

\bibitem[AI(2024{\natexlab{b}})]{anthropic_tracing_thoughts_2024}
AI, A.
\newblock Tracing thoughts in language models: Interpretability progress at anthropic.
\newblock \url{https://www.anthropic.com/research/tracing-thoughts-language-model}, 2024{\natexlab{b}}.

\bibitem[Ameisen et~al.(2025)Ameisen, Lindsey, Pearce, Gurnee, Turner, Chen, Citro, Abrahams, Carter, Hosmer, Marcus, Sklar, Templeton, Bricken, McDougall, Cunningham, Henighan, Jermyn, Jones, Persic, Qi, Ben~Thompson, Zimmerman, Rivoire, Conerly, Olah, and Batson]{ameisen2025circuit}
Ameisen, E., Lindsey, J., Pearce, A., Gurnee, W., Turner, N.~L., Chen, B., Citro, C., Abrahams, D., Carter, S., Hosmer, B., Marcus, J., Sklar, M., Templeton, A., Bricken, T., McDougall, C., Cunningham, H., Henighan, T., Jermyn, A., Jones, A., Persic, A., Qi, Z., Ben~Thompson, T., Zimmerman, S., Rivoire, K., Conerly, T., Olah, C., and Batson, J.
\newblock Circuit tracing: Revealing computational graphs in language models.
\newblock \emph{Transformer Circuits Thread}, 2025.
\newblock URL \url{https://transformer-circuits.pub/2025/attribution-graphs/methods.html}.

\bibitem[Amodei et~al.(2016)Amodei, Olah, Steinhardt, Christiano, Schulman, and Mané]{amodei2016concreteproblemsaisafety}
Amodei, D., Olah, C., Steinhardt, J., Christiano, P., Schulman, J., and Mané, D.
\newblock Concrete problems in ai safety, 2016.
\newblock URL \url{https://arxiv.org/abs/1606.06565}.

\bibitem[Anthropic(2025{\natexlab{a}})]{anthropic_biorisk_2025}
Anthropic.
\newblock Biological risk and model misuse.
\newblock \url{https://red.anthropic.com/2025/biorisk/}, 2025{\natexlab{a}}.
\newblock Accessed: 2025-10-18.

\bibitem[Anthropic(2025{\natexlab{b}})]{anthropic_cyber_2025}
Anthropic.
\newblock Cyber competitions and model misuse.
\newblock \url{https://red.anthropic.com/2025/cyber-competitions/}, 2025{\natexlab{b}}.
\newblock Accessed: 2025-10-18.

\bibitem[Bowen et~al.(2024)Bowen, Murphy, Cai, Khachaturov, Gleave, and Pelrine]{bowen2024scaling}
Bowen, D., Murphy, B., Cai, W., Khachaturov, D., Gleave, A., and Pelrine, K.
\newblock Scaling laws for data poisoning in llms.
\newblock \emph{arXiv e-prints}, pp.\  arXiv--2408, 2024.

\bibitem[Burns et~al.(2022)Burns, Ye, Klein, and Steinhardt]{burns2022discovering}
Burns, C., Ye, H., Klein, D., and Steinhardt, J.
\newblock Discovering latent knowledge in language models without supervision.
\newblock \emph{arXiv preprint arXiv:2212.03827}, 2022.

\bibitem[Cloud et~al.(2025)Cloud, Le, Chua, Betley, Sztyber-Betley, Hilton, Marks, and Evans]{cloud2025subliminallearninglanguagemodels}
Cloud, A., Le, M., Chua, J., Betley, J., Sztyber-Betley, A., Hilton, J., Marks, S., and Evans, O.
\newblock Subliminal learning: Language models transmit behavioral traits via hidden signals in data, 2025.
\newblock URL \url{https://arxiv.org/abs/2507.14805}.

\bibitem[Gloaguen et~al.(2025)Gloaguen, Vero, Staab, and Vechev]{gloaguen2025finetuning}
Gloaguen, T., Vero, M., Staab, R., and Vechev, M.
\newblock Finetuning-activated backdoors in llms.
\newblock \emph{arXiv preprint arXiv:2505.16567}, 2025.

\bibitem[Greenblatt et~al.(2024)Greenblatt, Denison, Wright, Roger, MacDiarmid, Marks, Treutlein, Belonax, Chen, Duvenaud, et~al.]{greenblatt2024alignment}
Greenblatt, R., Denison, C., Wright, B., Roger, F., MacDiarmid, M., Marks, S., Treutlein, J., Belonax, T., Chen, J., Duvenaud, D., et~al.
\newblock Alignment faking in large language models.
\newblock \emph{arXiv preprint arXiv:2412.14093}, 2024.

\bibitem[He et~al.(2024)He, Xia, and Henderson]{he2024your}
He, L., Xia, M., and Henderson, P.
\newblock What is in your safe data? identifying benign data that breaks safety.
\newblock \emph{arXiv preprint arXiv:2404.01099}, 2024.

\bibitem[Hubinger et~al.(2024)Hubinger, Denison, Mu, Lambert, Tong, MacDiarmid, Lanham, Ziegler, Maxwell, Cheng, et~al.]{hubinger2024sleeper}
Hubinger, E., Denison, C., Mu, J., Lambert, M., Tong, M., MacDiarmid, M., Lanham, T., Ziegler, D.~M., Maxwell, T., Cheng, N., et~al.
\newblock Sleeper agents: Training deceptive llms that persist through safety training.
\newblock \emph{arXiv preprint arXiv:2401.05566}, 2024.

\bibitem[Lee et~al.(2024)Lee, Phatale, Mansoor, Mesnard, Ferret, Lu, Bishop, Hall, Carbune, Rastogi, and Prakash]{lee2024rlaifvsrlhfscaling}
Lee, H., Phatale, S., Mansoor, H., Mesnard, T., Ferret, J., Lu, K., Bishop, C., Hall, E., Carbune, V., Rastogi, A., and Prakash, S.
\newblock Rlaif vs. rlhf: Scaling reinforcement learning from human feedback with ai feedback, 2024.
\newblock URL \url{https://arxiv.org/abs/2309.00267}.

\bibitem[Meng et~al.(2023)Meng, Bau, Andonian, and Belinkov]{meng2023locatingeditingfactualassociations}
Meng, K., Bau, D., Andonian, A., and Belinkov, Y.
\newblock Locating and editing factual associations in gpt, 2023.
\newblock URL \url{https://arxiv.org/abs/2202.05262}.

\bibitem[Ouyang et~al.(2022)Ouyang, Wu, Jiang, Almeida, Wainwright, Mishkin, Zhang, Agarwal, Slama, Ray, Schulman, Hilton, Kelton, Miller, Simens, Askell, Welinder, Christiano, Leike, and Lowe]{ouyang2022traininglanguagemodelsfollow}
Ouyang, L., Wu, J., Jiang, X., Almeida, D., Wainwright, C.~L., Mishkin, P., Zhang, C., Agarwal, S., Slama, K., Ray, A., Schulman, J., Hilton, J., Kelton, F., Miller, L., Simens, M., Askell, A., Welinder, P., Christiano, P., Leike, J., and Lowe, R.
\newblock Training language models to follow instructions with human feedback, 2022.
\newblock URL \url{https://arxiv.org/abs/2203.02155}.

\bibitem[{Psychiatrist.com}(2023)]{neda_chatbot_2023}
{Psychiatrist.com}.
\newblock Neda suspends ai chatbot for giving harmful eating disorder advice.
\newblock \url{https://www.psychiatrist.com/news/neda-suspends-ai-chatbot-for-giving-harmful-eating-disorder-advice/}, 2023.
\newblock Accessed: 2025-10-18.

\bibitem[Wei et~al.(2023)Wei, Huang, Lu, Zhou, and Le]{wei2023simple}
Wei, J., Huang, D., Lu, Y., Zhou, D., and Le, Q.~V.
\newblock Simple synthetic data reduces sycophancy in large language models.
\newblock \emph{arXiv preprint arXiv:2308.03958}, 2023.

\bibitem[Zou et~al.(2025)Zou, Phan, Chen, Campbell, Guo, Ren, Pan, Yin, Mazeika, Dombrowski, Goel, Li, Byun, Wang, Mallen, Basart, Koyejo, Song, Fredrikson, Kolter, and Hendrycks]{zou2025representationengineeringtopdownapproach}
Zou, A., Phan, L., Chen, S., Campbell, J., Guo, P., Ren, R., Pan, A., Yin, X., Mazeika, M., Dombrowski, A.-K., Goel, S., Li, N., Byun, M.~J., Wang, Z., Mallen, A., Basart, S., Koyejo, S., Song, D., Fredrikson, M., Kolter, J.~Z., and Hendrycks, D.
\newblock Representation engineering: A top-down approach to ai transparency, 2025.
\newblock URL \url{https://arxiv.org/abs/2310.01405}.

\end{thebibliography}

\newpage



\end{document}